\documentclass[11pt,a4paper]{article}
\usepackage[hyperref]{emnlp-ijcnlp-2019}
\usepackage{times}
\usepackage{latexsym}
\usepackage{times}
\usepackage{latexsym}
\usepackage{times}
\usepackage{tikz}

\def\checkmark{\tikz\fill[scale=0.4](0,.35) -- (.25,0) -- (1,.7) -- (.25,.15) -- cycle;} 
\usepackage[symbol]{footmisc}
\usepackage{float}

\usepackage{sfmath}
\usepackage{graphicx}
\usepackage{amsmath}
\usepackage{amssymb}
\usepackage[mode=buildnew]{standalone}
\usepackage{todonotes}
\usepackage{subfig}
\usepackage{url}
\usepackage{etoolbox}
\usepackage{diagbox}
\usepackage{amsmath}
\usepackage{algorithm}
\usepackage[noend]{algpseudocode}
\usepackage{soul}

\aclfinalcopy 

\newcommand*{\affaddr}[1]{#1} 
\newcommand*{\affmark}[1][*]{\textsuperscript{#1}}

\title{Talk2Car: Taking Control of Your Self-Driving Car}

\author{Thierry Deruyttere\affmark[1] \hspace{3mm} Simon Vandenhende\affmark[2] \hspace{3mm} Dusan Grujicic\affmark[2]\\\textbf{ \hspace{3mm} Luc Van Gool\affmark[2] \hspace{3mm} Marie-Francine Moens\affmark[1]} \\
\affaddr{\affmark[1]Department of Computer Science (CS)}\\
\affaddr{\affmark[2]Department of Electrical Engineering (ESAT)}\\
KU Leuven\\
{\tt\small \{thierry.deruyttere, sien.moens\}@cs.kuleuven.be}\\
{\tt\small \{simon.vandenhende, dusan.grujicic, luc.vangool\}@esat.kuleuven.be}}

\date{}
\begin{document}
\maketitle
\begin{abstract}
A long-term goal of artificial intelligence is to have an agent execute commands communicated through natural language.
In many cases the commands are grounded in a visual environment shared by the human who gives the command and the agent. Execution of the command then requires mapping the command into the physical visual space,  
after which the appropriate action can be taken.
In this paper we consider the former. 
Or more specifically, we consider the problem in an autonomous driving setting, where a passenger requests an action that can be associated with an object found in a street scene.
Our work presents the \texttt{Talk2Car} dataset, which is the first object referral dataset that contains commands written in natural language for self-driving cars.
We provide a detailed comparison with related datasets such as \texttt{ReferIt}, \texttt{RefCOCO}, \texttt{RefCOCO+}, \texttt{RefCOCOg}, \texttt{Cityscape-Ref} and \texttt{CLEVR-Ref}.
Additionally, we include a performance analysis using strong state-of-the-art models.
The results show that the proposed object referral task is a challenging one for which the models show promising results but still require additional research in natural language processing, computer vision and the intersection of these fields. 
The dataset can be found on our website: \hiddenurl{http://macchina-ai.eu/} 
\end{abstract}

\section{Introduction}
Researchers have studied the problem of understanding actions communicated through natural language in both simulated \cite{Das2017, IQA, Hermann2017} and real environments \cite{arid, TalkTheWalk, Anderson2017}.
This paper focuses on the latter.
More concretely, we consider the problem in an autonomous driving setting, where a passenger can control the actions of an \textbf{A}utonomous \textbf{V}ehicle (AV) by giving natural language commands.
We hereunder argue why this problem setting is particularly interesting. 

\begin{figure*}
\begin{center}
 \centering
\subfloat[You can park up ahead behind \textbf{the silver car}, next to that lamppost with the orange sign on it]{\includegraphics[width=.32\linewidth]{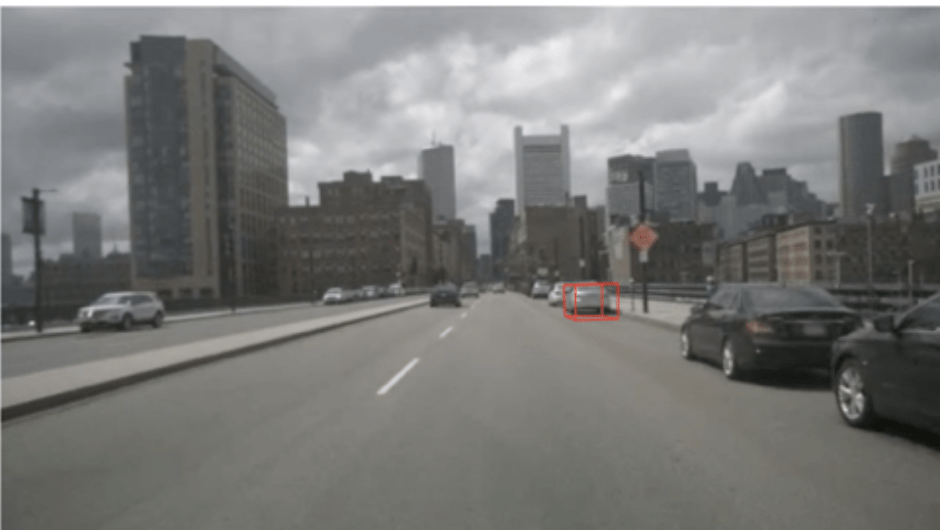}}
\hfill
\subfloat[\textbf{My friend} is getting out of the car. That means we arrived at our destination! Stop and let me out too!]{\includegraphics[width=.32\linewidth]{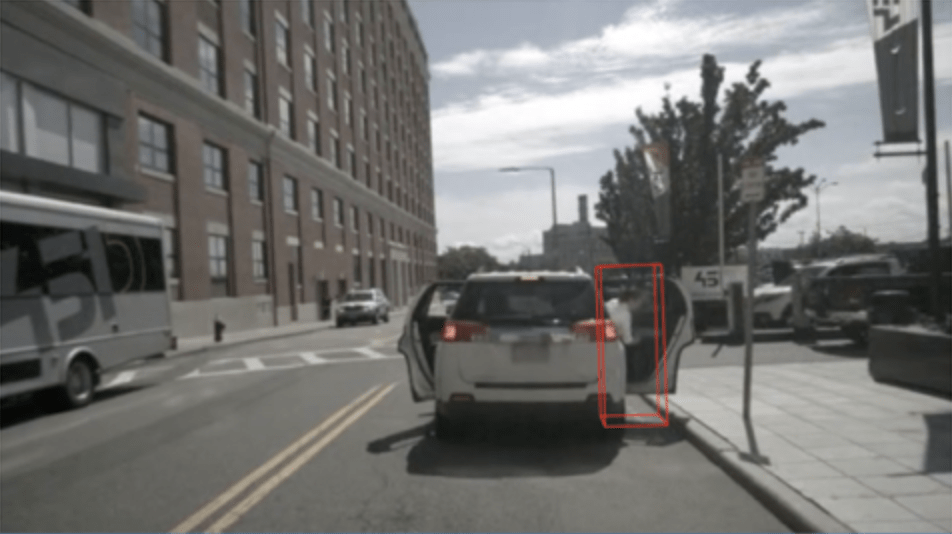}}
    \hfill
\subfloat[Yeah that would be \textbf{my son} on the stairs next to the bus. Pick him up please]{\includegraphics[width=.32\linewidth]{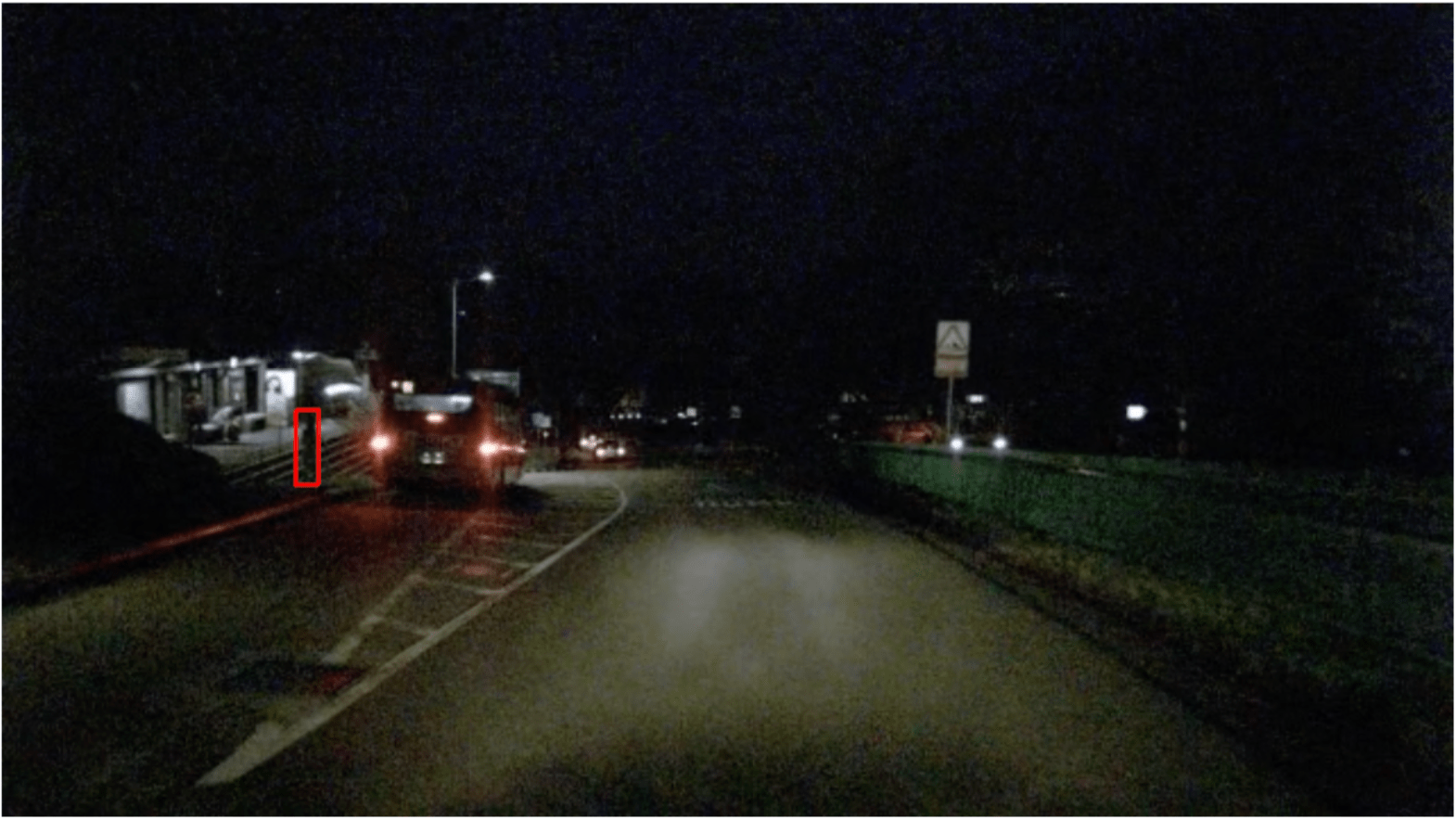}}
\hfill 
\subfloat[After \textbf{that man in the blue top} has passed, turn left]{\includegraphics[width=.32\linewidth]{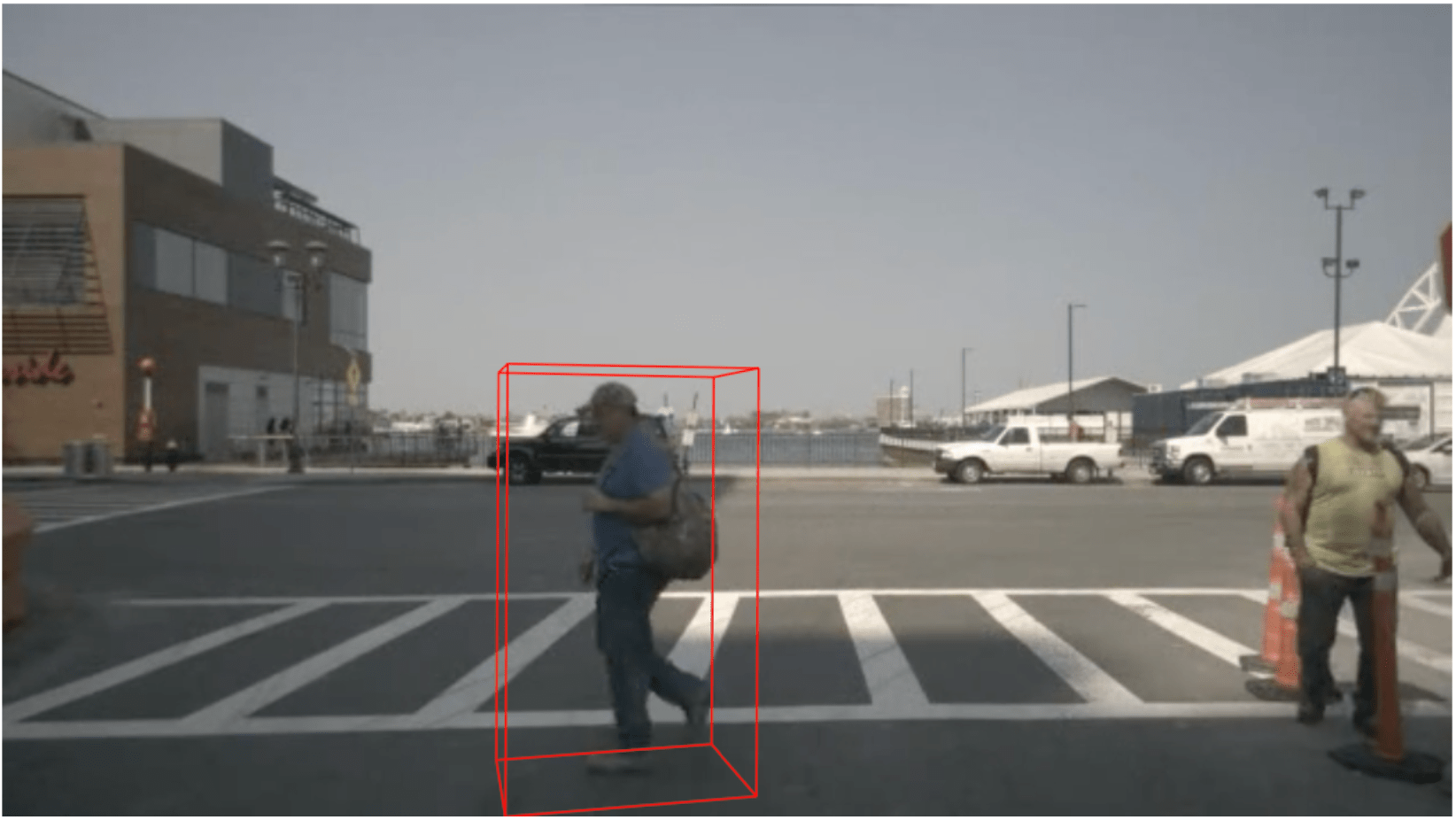}}
    \hfill
\subfloat[There's \textbf{my mum}, on the right! The one walking closest to us. Park near \textbf{her}, she might want a lift]{\includegraphics[width=.32\linewidth]{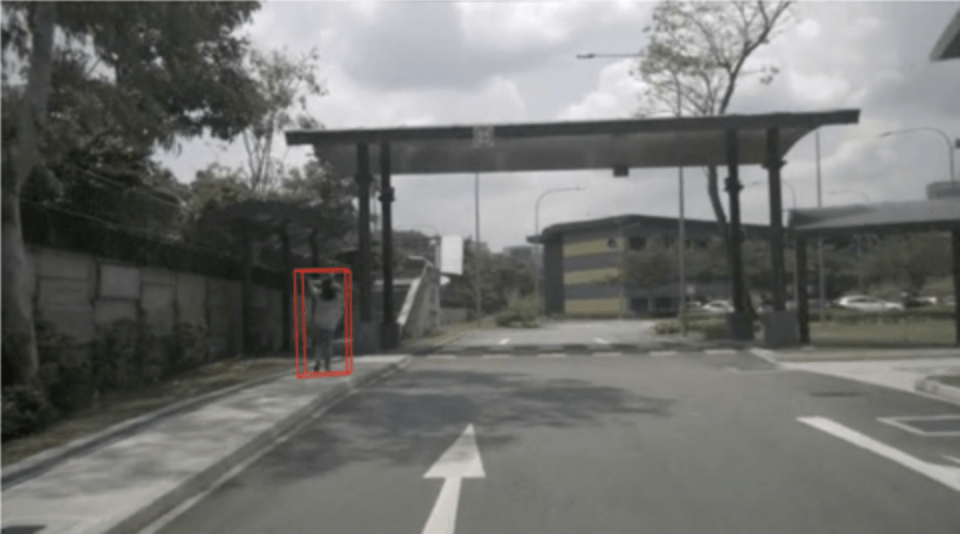}}
\hfill
\subfloat[Turn around and park in front of \textbf{that vehicle in the shade}]{\includegraphics[width=.32\linewidth]{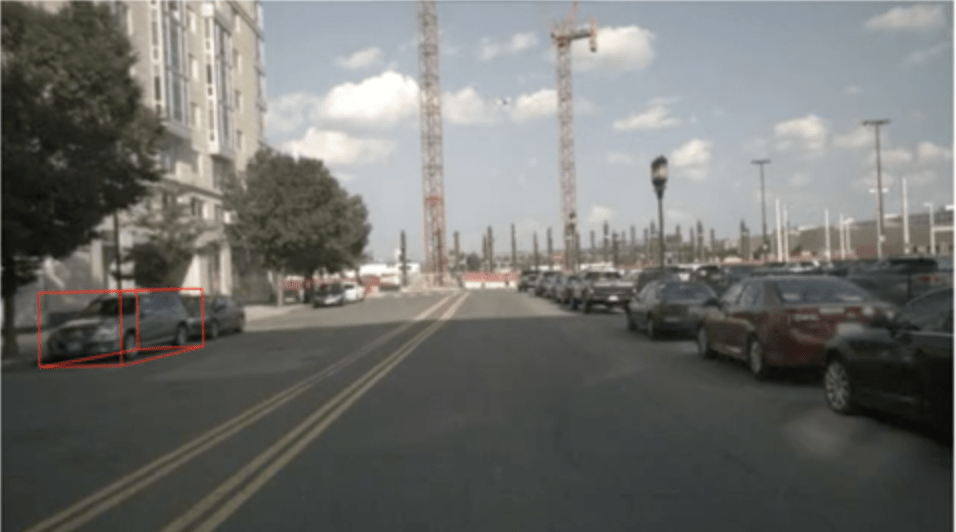}}
\end{center}
   \caption{The \texttt{Talk2Car} dataset adds textual annotations on top of the \texttt{nuScenes} dataset for urban scene understanding. The textual annotations are free form commands, which guide the path of an autonomous vehicle in the scene. Each command describes a change of direction, relevant to a referred object found in the scene (here indicated by the red 3D-bounding box). Best seen in color.}
\label{fig:exampleCommand}
\end{figure*}

\paragraph{}
First, a recent study by \newcite{PublicPerceptionSDC} 
has shown that the majority of the public is reluctant to step inside an AV.
A possible explanation for this might be the lack of control which can be unsettling to some.
Providing a way to communicate with the vehicle could help alleviate this uneasiness. 
Second, an AV can become hesitant in some situations \cite{self-driving-car-problems}.
By giving a task or command, the passenger could guide the agent in its decision process.
Third, some situations request feedback.
For example, a passenger might 
indicate that they want to park in the shade during a sunny day.
Finally, the problem of urban scene understanding is one of practical relevance that has been well studied \cite{CityScape,geiger2013vision}.
We believe all of this makes it an interesting setting to assess the performance of grounding natural language commands into the visual space.
\newline

To perform the requested action, an agent is required to take two steps.
First, the agent needs to interpret the command and ground it into the physical visual space.
Secondly, the agent has to devise a plan to execute the given command.
This paper focuses on this former step, or more concretely:
given an image $I$ and a command $C$, the goal is to find the region $R$ in the image $I$ that the command is referring to.
In this paper, to reduce the complexity of the object referral task we restrict the task to the case where there is only one targeted object that is referred to in the natural language command.

To stimulate research on grounding commands into the visual space we present the first object referral dataset, named \texttt{Talk2Car}, 
that comes with commands formulated in textual natural language for self-driving cars. 
A few example commands together with their contextual images can be found in Fig. \ref{fig:exampleCommand}.
Moreover, by using this new dataset we evaluate the performance of several strong state-of-the-art models that recognize the referred object of a command in the visual scene. Here we encounter several challenges. Referred objects are sometimes ambiguous (e.g., there are several cyclists in the scene), but can be disambiguated by understanding modifier expressions in language (e.g., the biker with the red jacket).
These modifier expressions could also indicate spatial information.
Furthermore, detecting the targeted object is challenging both in the language utterance and the urban scene, for instance, when dealing with complex and long sentences which might contain coreferent phrases, and with distant objects in the visual scene, respectively.
Finally, in AV settings the speed of predicting the location of the referred object is of primordial importance.  

The contributions of our work are the following:
\begin{itemize}
    \item{We propose the first object referral dataset for grounding commands for self-driving cars in free natural language into the visual context of a city environment.}
    \item{We evaluate several state-of-the-art models that recognize the referred object of a natural language command in the urban scene}.
     \item{We especially evaluate the models 1) for their capabilities to disambiguate objects based on modifying and spatial relationships expressed in language; 2)
     for their capabilities to cope with difficult language and visual context; and 3) with respect to prediction speed, which is important in real-life AV settings.}
\end{itemize}
\section{Related Work}
\label{sect:related-work}

\paragraph{Object Referral}
The \texttt{Talk2Car} dataset considers the object referral task, which requires to retrieve the correct object (region) from an image based on a language expression.
A common method is to first extract regions of interest from the image, using a region proposal network (RPN). 
\citet{yu2016modeling, mao2016generation} decode these proposals as a caption using a recurrent neural network (RNN).
The predicted region corresponds to the caption that is ranked most similar to the referring expression.
Other works based on RPN \cite{hu2017modeling} or Faster-RCNN \cite{yu2018mattnet} have integrated attention mechanisms to decompose the language expressions into multiple sub-parts but use tailored modules for specific sub-tasks making them less fit for our object referral task. 
\citet{karpathy2014deep} interpret the inner product between region proposals and sentence fragments as a similarity score, allowing to match them in a bidirectional manner. 
\citet{hu2016natural} uses an encoding of the global context in addition to the local context from the extracted regions. 
\citet{STACK} explore the use of modular networks for this task.
They are comprised of multiple smaller predefined building blocks that can be combined together based on the language expression.
The 
last three state-of-the-art models are evaluated on \texttt{Talk2Car} (section \ref{sect:sota}).


\paragraph{Grounding in Human-Robot Interaction}
When giving commands to robots, the grounding of the command in the visual environment is an essential task.  
\citet{deits2013clarifying} 
use Generalized Grounding Graphs ($G^3$) \cite{tellex2011understanding, kollar2013generalized} which is a probabilistic graphical model based on the compositional and hierarchical structure of a natural language command. 
This approach allows to ground certain parts of an image with linguistic constituents.
\citet{shridhar2018interactive} consider the task where a robot arm has to pick up a certain object based on a given command.
This is accomplished by creating captions for extracted regions from a RPN and clustering them together with the original command. If the command is ambiguous and more than one caption indicates the 
referring expression, the system will ask a clarifying question in order to be able to pick the right object. Due to its computational complexity during prediction, we did not select the last model in our evaluations.  

\paragraph{Visual Question Answering}
The goal of 
VQA is to ask any type of question about an image for which the system should return the correct answer.
This requires the system to have a good understanding of the image and the question.
Early work 
\cite{kafle2016answer, ZhouTSSF15, FukuiPYRDR16}
tried to solve the task by fusing image features extracted by a convolutional neural network (CNN), together with an encoding of the question. 
\cite{Johnson2017, Suarez2018} experimented with modular networks for this task. 
\citet{Hudson2018} proposed the use of a network made of recurrent Memory, Attention and Composition (MAC) cells. Similar to modular networks, the MAC model also uses multiple reasoning steps, making it a suitable model in our evaluation (section  
\ref{sect:sota}).

\paragraph{Object Referral Datasets}
Over the years, various object referral datasets based on both real world and computer generated images have been proposed.
\citet{ReferIt} introduced the first real-world large-scale object referral dataset named \texttt{ReferIt}.
\citet{yu2016modeling} constructed \texttt{RefCOCO} and \texttt{RefCOCO+} and as the names suggest, these two datasets are based on the \texttt{MSCOCO} dataset \cite{MSCOCO}.
A third dataset also based on \texttt{MSCOCO}, named \texttt{RefCOCOg} \cite{mao2016generation}, contains longer language expressions than the previous two datasets. 
This dataset has an average expression length of 8.43 words per expression compared to 3.61 for \texttt{RefCOCO} and 3.53 for \texttt{RefCOCO+}. 
The dataset closest to ours is the dataset by \citet{vasudevan2018object}, as it augments \texttt{Cityscapes} \cite{CityScape} with textual annotations. We will henceforth refer to this dataset as \texttt{Cityscapes-Ref}. The main difference with this work, is that the \texttt{Talk2Car} dataset contains commands, rather than descriptions. 
A computer generated dataset named \texttt{CLEVR-Ref} was proposed by \citet{STACK} which has been created by augmenting the \texttt{CLEVR} dataset \cite{CLEVR} such that it would include referred objects. 
We refer to section \ref{sect:compare-datasets} for a thorough comparison between these datasets and ours.
Some new datasets have recently been proposed where a car has to navigate through a city based on a textual itinerary given by a passenger and locate a target at the final destination
\cite{chen2019touchdown, vasudevan2019talk2nav}. 
This can also be seen as an object referral task with the addition of following an itinerary.
While being a very interesting problem, it differs from the task being evaluated in this paper

\section{Dataset}
\subsection{Dataset Collection and Annotation}
\label{subsect:dataset-collection}
The \texttt{Talk2Car} dataset is built upon the \texttt{nuScenes} dataset \cite{nuScenes} which is a large-scale dataset for autonomous driving. 
The \texttt{nuScenes} dataset contains 1000 videos of 20 seconds each taken in different cities (Boston and Singapore), weather conditions (rain and sun) and different times of day (night and day).
These videos account for a total of approximately 1.4 million images. Each scene comes with data from six cameras placed at different angles on the car, LIDAR, GPS, IMU, RADAR and 3D bounding box annotations. The 3D bounding boxes discriminate between 23 different object classes. 
\begin{figure*}[ht]
\begin{center}
\includestandalone[width=0.7\linewidth]{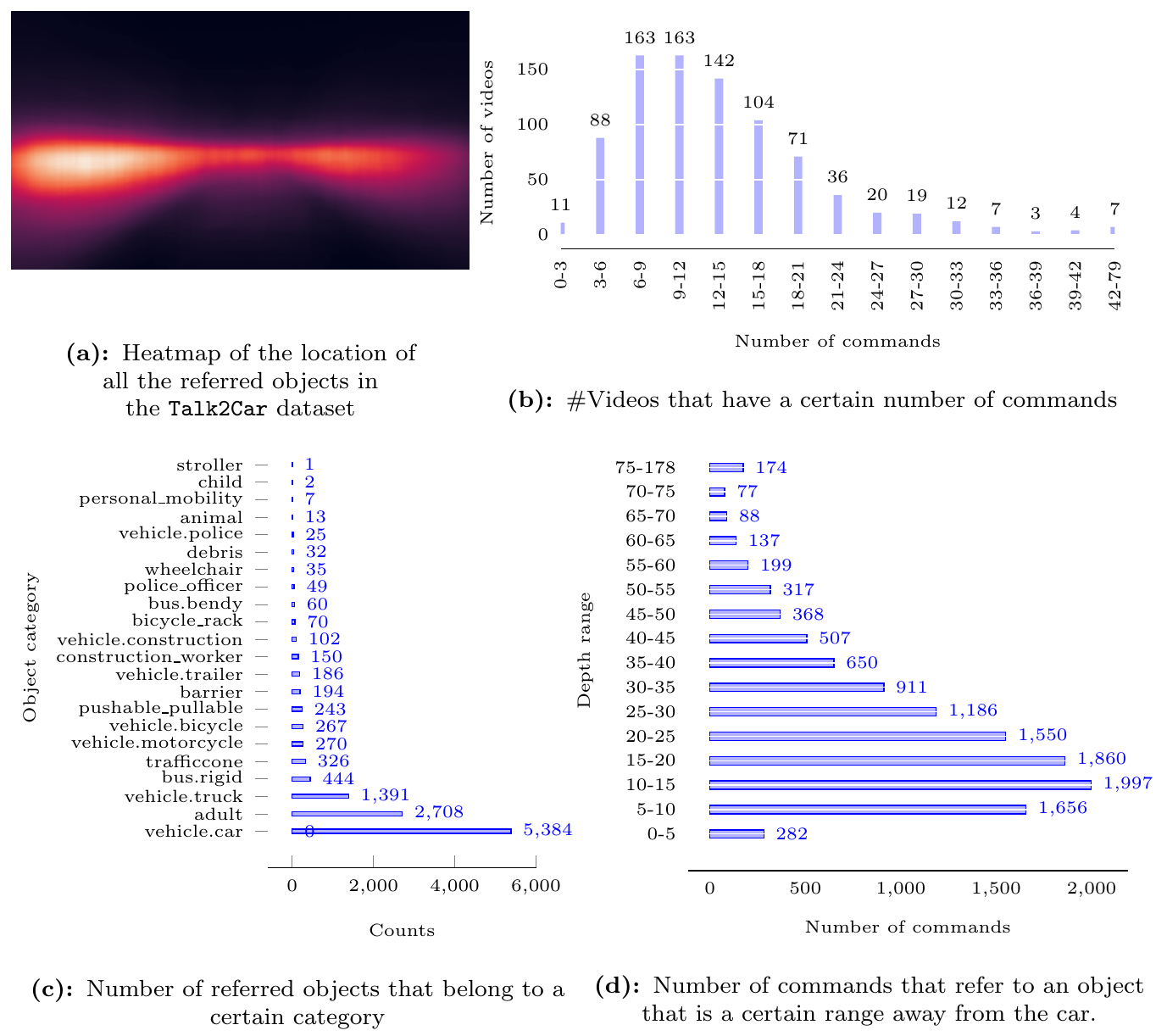}
\end{center}
\caption{Statistics of the \texttt{Talk2Car} dataset.}
\label{fig:stats}
\end{figure*}

We relied on workers of Amazon Mechanical Turk (AMT) to extend the videos from the \texttt{nuScenes} dataset with written commands. 
To create commands, each worker watches an entire 20 second long video from the front facing camera. Afterwards, the worker navigates to any point in the video that is found interesting. Once the worker has decided on the frame, a pre-annotated object from the \texttt{nuScenes} dataset for that frame needs to be selected. 
The annotation task is completed when a command referring to the selected object is entered. The workers were free to enter any command, as long as the car can follow a path based on the command.\footnote{
The path to be followed by the car when executing the command, will be added in a later version of the dataset.} 

We hired five workers per video who could enter as many commands per video frame as they wanted. To ensure high quality annotations, we manually verified the correctness of all the commands and corresponding bounding boxes. The verification happened in a two-round system, where each annotation had to be qualified as adequate by two different reviewers. To incentivize workers to come up with diverse and meaningful commands, we awarded a bonus every time their work received approval. 

\subsection{Statistics of the Dataset}
\label{subsect:statistics}
The \texttt{Talk2Car} dataset contains 11 959 commands for the 850 videos of the \texttt{nuScenes} training set as 3D bounding box annotations for the test set of the latter dataset are not disclosed.
55.94\% and 44.06\% of these commands belong to videos taken respectively in Boston and Singapore.
On average a command consist of 11.01 words, 2.32 nouns, 2.29 verbs and 0.62 adjectives.
Each video has on average 14.07 commands.
In Fig. \ref{fig:stats}(d) we can see the distribution of distance to the referred objects.
Fig. \ref{fig:stats}(b) displays the distribution of commands over the videos. 
On average there are 4.27 objects with the same category as the referred object per image and on average there are 10.70 objects per image.
Fig. \ref{fig:stats}(a) shows a heatmap of the location of all referred objects in the images of \texttt{Talk2Car}. 
In Fig. \ref{fig:stats}(c) we see the distribution of commands that refer to an object of a certain category.

\subsection{Dataset Splits}
\label{subsect:splits}
We have split the dataset in such a way that the train, validation and test set would contain 70\%, 10\% and 20\% of the samples, respectively. 
To ensure a proper coverage of the data distribution in each set, we have taken a number of constraints into account.
First, samples belonging to the same video are part of the same set. 
Second, as the videos are shot in either Singapore or Boston, i.e., in left or right hand traffic, the distributions of every split have to reflect this. 
Third, we aim to have a similar distribution of scene conditions across different sets, such as the type of weather and the time of day.
Finally, as the number of occurrences of object categories is heavily imbalanced 
(see fig. \ref{fig:stats}(c)), 
we have ensured that every object category contained in the test set is also present in the training set. 
With these constraints in mind we randomly sample the three sets for 10 000 times and optimize for a data distribution of 70\%, 10\% and 20\%.
The resulting train, validation and test sets contain 8 349 (69.8\%), 1 163 (9.7\%) and 2 447 (20.4\%) commands respectively. 
We have also identified multiple subsets of the test set, which allow evaluation of specific situations. When the referred object category occurs multiple times in an image, 
attributes in modifying expressions in language including spatial expressions might disambiguate the referred object. This has led to test sets with different numbers of occurrences of the targeted category of the referred object. Longer commands might contain irrelevant information or might be more complex to understand, leading to test sets with commands of different length. Finally, referred objects at large distances from the AV might be difficult to recognize. Hence, we have built test sets that contain referred objects at different distances from the AV.
\section{Datasets Comparisons}
\subsection{Quantitative Evaluation of \texttt{Talk2Car}}
\label{sect:compare-datasets}


Table \ref{tab:comparison} compares the \texttt{Talk2Car} dataset with prior object referral datasets. 
It can be seen that \texttt{Talk2Car}
contains fewer natural language expressions than the others.
However, although the dataset is smaller, the expressions are of high quality thanks to the double review system we discussed earlier in section \ref{subsect:dataset-collection}.
The main reason for having fewer annotations is that the original \texttt{nuScenes} dataset only discriminates between 23 different categories corresponding to the annotated bounding boxes.
Moreover, the original \texttt{nuScenes} dataset considers the specific setting of urban scene understanding.
This limits the visual domain considerably in comparison to \texttt{MS-COCO}.
On the other hand, \texttt{Talk2Car} contains images in realistic settings accompanied by free language in contrast to curated datasets such as MS-COCO.
Compared to most of the above datasets, the video frames annotated with natural language commands are part of larger videos that contain in total 1 183 790 images which could be exploited in the object referral task. 

When we consider the average length of the natural language expressions in \texttt{Talk2Car}, we find that it ranks third, after \texttt{CLEVR-Ref} and \texttt{Cityscapes-Ref}. 
We did not put limitations on what the language commands could contain, which benefits 
the complexity and linguistic diversity of the expressions (section \ref{subsect:qualitative-evaluation}). 

When looking at the type of modalities, the \texttt{Talk2Car} dataset considers RADAR, LIDAR and video. 
These modalities are missing in prior work except for video in \texttt{Cityscapes-Ref}. 
Including various modalities allows researchers to study a very broad range of topics with just a single dataset. 

\begin{table*}[ht!]
\small
\begin{center}
\scalebox{0.9}{
\begin{tabular}{|l|c|c|c|c|c|c|c|c|}
\hline
Dataset & Images & Objects & Expressions & Avg expr length  & Video & Lidar & Radar \\
\hline
\texttt{ReferIt} \cite{ReferIt} &  19,894 &  96,654  & 130,525 & 3.46 &$\times$&$\times$&$\times$ \\
\texttt{RefCOCO} \cite{yu2016modeling} &  26,711 & 50,000 & 142,209 & 3.61 &$\times$&$\times$&$\times$ \\
\texttt{RefCOCO+} \cite{yu2016modeling} & 19,992 & 49,856 & 141,564 & 3.53 &$\times$&$\times$&$\times$ \\
\texttt{RefCOCOg} \cite{mao2016generation}& 26,711 & 54,822 & 85,474  & 8.43 &$\times$&$\times$&$\times$ \\
\texttt{CLEVR-Ref} \cite{STACK} & 99,992 & 492,727 & 998,743 & 14.50 &$\times$&$\times$&$\times$  \\ 
\texttt{Cityscapes-Ref} \cite{vasudevan2018object} & 4,818 &  29,901  & 30,000  & 15.59 &\checkmark& $\times$ & $\times$ \\
\hline
\textbf{\texttt{Talk2Car} (Ours)} & 9,217 & 10,519 & 11,959 & 11.01 &\checkmark &\checkmark &\checkmark  \\
\hline
\end{tabular}
}
\end{center}
\caption{Statistics of and comparison with existing datasets for object referral.}
\label{tab:comparison}
\end{table*}
\vspace{-.01cm}
\subsection{Qualitative Evaluation of \texttt{Talk2Car}}
\label{subsect:qualitative-evaluation}
To make our discussion more concrete, we compare the textual annotations from Fig. \ref{fig:exampleCommand} with some examples from prior work that are listed below.
\texttt{RefCOCO} contains expressions such as `Woman on right in white shirt' or `Woman on right'. 
\texttt{RefCOCO+} on the other hand contains expressions such as `Guy in yellow dribbling ball' or `Yellow shirt in focus'.
Lastly, \texttt{ReferIt} contains `Right rocks', `Rocks along the right side'.
The language used in the above prior work is more simple, explicit and is well structured in comparison to the commands of \texttt{Talk2Car}. 
Additionally, the latter
tend to include irrelevant side-information, e.g., 'She might want a lift', instead of being merely descriptive.
The unconstrained free language of \texttt{Talk2Car} introduces different challenges, which involve
co-reference resolution, named entity recognition, understanding relationships between objects, linking attributes to objects and understanding which object is the object of interest in the command.

The commands also contain implicit referrals as can be seen 
in the command in Fig. \ref{fig:exampleCommand}(f):
`Turn around and park in front of that vehicle in the shade'. 
Similar to 
\texttt{CLEVR-Ref}, object referral in \texttt{Talk2Car} requires some form of spatial reasoning.
However, in contrast to the former, there are cases where the spatial description in the command is misleading and truthfully reflects mistakes that people make.
An example is the command in Fig. \ref{fig:exampleCommand}(e), where we refer to the object as being on the right side of the image, while the person of interest is actually located on the left. 

Another important difference is the type of images in each dataset. For instance, the urban images in RefCOCO are taken from the viewpoint of a pedestrian. On the other hand, the images in Talk2Car are car centric.
\section{Application of the State-of-the-Art Models and their Evaluation}
\label{sect:sota}
We assess the performance of 7 models to detect the referred object in a command on the \texttt{Talk2Car} dataset. 
We discriminate between state-of-the-art methods based on region proposals and non-region proposal methods, apart from simple baselines. \footnote {All parameter values obtained on the validation set are cited in the supplementary material. For both the MAC and STACK-NMN we give the best results after empirically setting the number of reasoning steps. }
\subsection{Region Proposal Based Methods}
\label{subsec: region_proposal_methods}
\paragraph{Object Sentence Mapping (OSM)}
This region proposal based method uses a single-shot detection model, i.e., SSD-512 \cite{liu2016ssd}, to extract 64 interest regions from the image. 
We pretrain the region proposal network (RPN) for the object detection task on the train images from the \texttt{Talk2Car} dataset.
A ResNet-18 model is used to extract a local representation for the proposed regions.
The natural language command is encoded using a neural network with Gated Recurrent Units (GRUs).
Inspired by \cite{karpathy2014deep}, we use the inner product between the latent representation of the region and command as a score for each proposal. 
The region that gets assigned the highest score is returned as bounding box for the object referred to by the command.

\paragraph{Spatial Context Recurrent ConvNet (SCRC)}
A shortcoming of the above baseline model is that the correct region has to be selected based on local information alone. Spatial Context Recurrent ConvNets \cite{hu2016natural} match both local and global information with an encoding of the command. We reuse the SSD-512 model from above to generate region proposals. A global image representation is extracted by a ResNet-18 model. Additionally, we add an 8-dimensional representation of the spatial configuration to the local representation of each bounding box, $x_{spatial} = \left[ x_{min}, y_{min}, x_{max}, y_{max}, h, w, x_{center}, y_{center} \right]$ with $h$ and $w$ respectively being the height and the width of this bounding box. For more details, we refer to the original work \cite{hu2016natural}.

\subsection{Non-Region Proposal Based Methods}
\paragraph{MAC model} This model \cite{Hudson2018} originally created for the VQA task uses a recurrent MAC cell to match the 
natural language command represented with a Bi-LSTM model with a global representation of the image. A ResNet-101 is used to extract the visual features from the image.
The MAC cell decomposes the textual input into a series of reasoning steps, where 
the MAC cell attends to certain parts of the textual input to guide the model to look at certain parts of the image. 
Between each of these reasoning steps, information is passed to the next cell such that the model is capable of representing arbitrarily complex 
reasoning graphs in a soft manner in a sequential way.
The recurrent control state of the MAC cell identifies a series of read and write operations.
The read 
unit extracts relevant information from both a given image and the internal memory. The 
write
unit iteratively integrates the information into the cells' memory state, producing a new intermediate result.

\begin{table*}[t]
\small
\begin{center}
\begin{tabular}{|l|c|c|c|}
\hline
Method & $IoU_{0.50}$ (\%) & Inference Speed (ms) & Params (M) \\
\hline
RS & 2.05 & - & - \\
BOBB & 3.15 & - & - \\ 
RNM & 9.70 & - & - \\
MAC \cite{Hudson2018}
& \textbf{50.51} & 51
& 41.59 \\
STACK-NMN \cite{STACK}
& 33.71 & 52
& 35.2 \\
OSM \cite{karpathy2014deep} & 35.31 & 71
& 43.0 \\
SCRC \cite{hu2016natural} & 38.70 & 90 & 59.5 \\
\hline
\end{tabular}
\end{center}
\caption{Performance ($IoU_{0.50}$), inference speed (evaluated on a TITAN XP) and number of parameters of the different models.}
\label{tab:experiments}
\end{table*}
\paragraph{Stack-NMN}

The Stack Neural Module Network or Stack-NMN \cite{STACK} uses 
multiple modules that can solve a task by automatically inducing a sub-task decomposition, where each sub-task is addressed by a separate neural module.
These modules can be chained together to decompose the natural language command into a reasoning process.
Like the MAC model, this reasoning step is based on the use of an attention mechanism to attend to certain parts of the natural language command, which on their turn guide the selection of neural modules.
The modules are first conditioned with the attended textual features after which they perform sub-task recognitions on the visual features.
The output of these modules are attended parts in the image which are then given to the next reasoning step to continue the reasoning process.
Again, a ResNet-101 model is used to extract the image features and a Bi-LSTM to encode the natural language command.
To predict the referred object this model first splits the given image into a 2D grid.
Then it tries to predict in which cell located in the grid the center of the referred object lies. 
Once this has been predicted, the model predicts the offsets of the bounding box relative to the predicted center.


\subsection{Simple Baselines}
\paragraph{Random Selection (RS)}
We reuse the single-shot detection model from section \ref{subsec: region_proposal_methods} to generate 64 region proposals per image of the test set. This model randomly samples one region from the proposals and uses it as prediction for the referred object. This is done 100 times and results are averaged.

\paragraph{Biggest Overlapping Bounding Box (BOBB)}
From the heatmap in Fig. \ref{fig:stats} (a) we can see that there is some bias of the referred objects on the left side. 
This model tries to exploit this information by searching a 2D bounding box that optimizes the overlap with all the bounding boxes in the training set.
The algorithm is explained in Section \ref{sec:supplemental} of the supplementary material.

\paragraph{Random Noun Matching (RNM)}
In the test set a dependency parser \cite{honnibal-johnson:2015:EMNLP} is used to extract the set of nouns from a given command. We keep the nouns which are substrings of the category names. Then, we randomly sample an object from the region proposals of the corresponding image. 
If the set of category names is empty, 
we randomly sample a region from all 
region proposals. 
We re-use the RPN explained in OSM for the region proposals.
This method is evaluated 100 times before averaging the results.
\subsection{Results and Discussion}

\label{sect:experiments}
\begin{figure*}[t!]
\begin{center}
\includestandalone[width=1\linewidth]{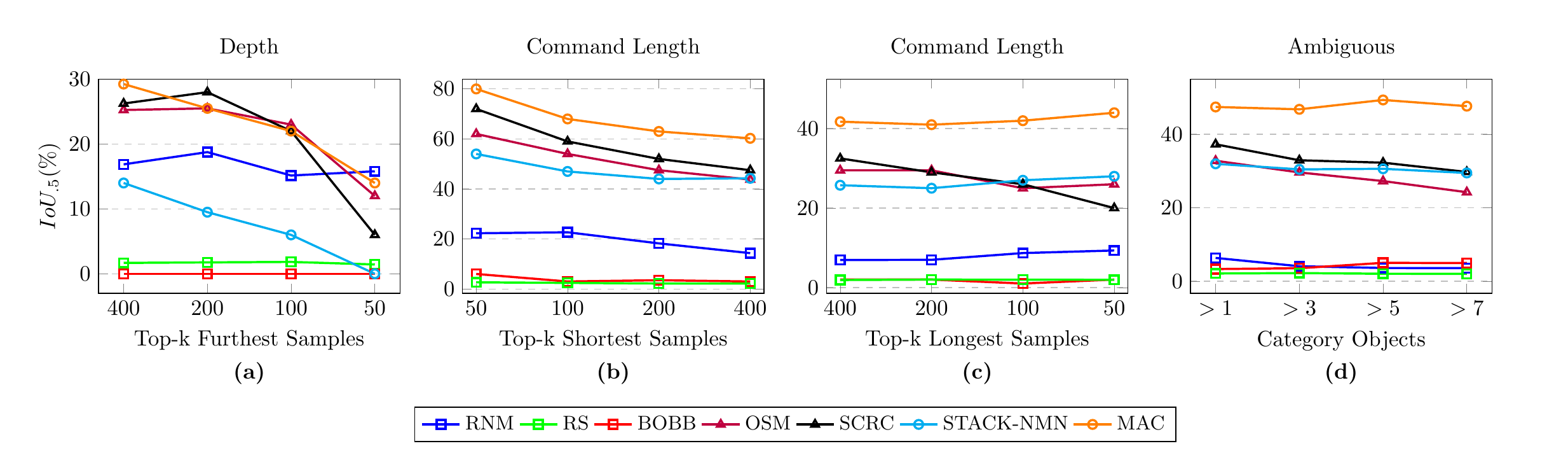}
\caption{Test performance $IoU_{0.5}$
of different methods on the challenging sub-test sets.
We discriminate between 
A test set for the top-k furthest objects in Fig. (\textbf{a}), the top-k shortest and longest commands in Fig. (\textbf{b}) and Fig. (\textbf{c}) respectively, and in function of the number of objects of the same category in the scene in Fig. (\textbf{d}).} 
\label{fig:subset_tests}
\end{center}
\end{figure*}

\paragraph{Overall Results} We evaluated all seven models on the object referral task, using both the test split from subsection \ref{subsect:splits} as well as multiple increasingly challenging subsets from this test set.
To properly evaluate existing models against our baselines we convert the 3D bounding boxes to 2D bounding boxes.
We consider the predicted region correct when the Intersection over Union (IoU), that is, the intersection of the predicted region and the ground truth region over the union of these two, is larger than 0.5. 
Additionally, we report average inference speed at prediction time per sample and number of parameters of each model. 
We report the results obtained on the test set in Table \ref{tab:experiments}. The results over the challenging test subsets can be seen in Fig. \ref{fig:subset_tests}.

In all results we see the following:
First, it is clear that the simple baselines (RS, BOBB, RNM) do not perform well, which evidences the difficulty of the object referral task in the realistic settings captured in \texttt{Talk2Car}.
Second, MAC performs the best on nearly all tasks and it performs significantly better than STACK-NMN which is the model that resembles MAC the most.

If we compare the two RPN systems
we see that SCRC often outperforms OSM, showing that using spatial information is beneficial. 
Third, being able to discriminate between the different object classes in the scene is important. Or put it differently, correct alignment between objects in the image and the category names mentioned in the command is a basic requirement. RNM shows us that concentrating on nouns already gives a big improvement over a purely random strategy. In a separate experiment using ground truth bounding boxes the RNM system obtained an IoU of 54\% showing the importance of the alignment of a found object to the correct category name.
Fourth, the command length has a negative impact on most models as can be seen in \ref{fig:subset_tests}(c). 
We argue that when commands get longer there might be more irrelevant information included which the models have difficulty to cope with.
Fifth, 
from our experiments we found that the non-RPN systems are roughly two times faster 
than the RPN-systems. 
This is due to the fact that these RPN-systems have to align every proposed region with the command.
On the other hand, the non-RPN systems only have to encode the full image once and then reason over this embedding. 
Lastly, when looking at the ambiguity test in Fig. \ref{fig:subset_tests}(d) we see that all models struggle when the number of ambiguous objects of one category increase except for STACK and MAC, whose performance remains fairly stable.
We believe they benefit from 
the multiple reasoning steps before giving an answer where modifier constructions in language disambiguate the referred object.
In a separate experiment we have focused on object referral in extra long commands with ambiguous objects of the same category, where we observe the same trends.
\vspace{-0.18cm}
\paragraph{Influence of Region Proposal Quality}
We consider the case when we pre-train a RPN on all keyframes from the training videos, rather than only on the images with commands. 
It is found that the test performance of the OSM model increased from 35.31 to 40.78\%.
Similarly, the test performance of the SCRC model increased from 38.70 to 41.15 \% showing the importance of starting from good region proposals. 


\paragraph{Blanking out the commands}
\cite{cirik2018visual} found that some referential datasets have some kind of bias in the dataset when blanking out the question. 
We evaluated this with both SCRC and OSM by changing the question vector to a zero filled vector and we respectively got the following results. For SCRC we get 40.37\% $IoU_0.5$ (38.70\% with command), OSM: 21.65\% (35.31\% with command). From these results we can conclude two things. First, global information which was added to the local representation of each region in the SCRC model, contains some kind of bias that the models can learn. Second, if no global information is used, as is the case in OSM, the model $IoU_0.5$ actually decreases dramatically indicating that there is not a high bias in the image itself.

\paragraph{Influence of Using Pre-trained Word Embeddings}
Using pre-trained word GloVe embeddings \cite{pennington2014glove}
had no effect on or even lowered the IoU obtained on the test set.
We argue that words like 'car' and 'truck' are very close to each other in the embedding space but for the model to perform well it should be able to discriminate between them.
We also tested ELMO \cite{Peters:2018} and BERT \cite{devlin2018bert} embeddings but found that they gave only minor improvements for some models.
\section{Conclusions and Future Work}
\label{sect:conclusion}
We have presented a new dataset, \texttt{Talk2Car}, that contains commands in natural language referring to objects in a visual urban scene which both the passenger and the self-driving car can see. 
We have compared this dataset to existing datasets for the joint processing of language and visual data and have performed experiments with different strong state-of-the-art models for object referral, which yielded promising results.
The available 3D information was neglected to be able to compare existing models but we believe that it 
could help in object referral as it contains more spatial information which, as seen in the experiments, is an important factor.
This 3D information will help to translate language into 3D. Moreover, it will allow to perform actions in 3D based on the given command.
Also, the \texttt{Talk2Car} dataset only allows people to refer to one object at a time. It also doesn't include path annotations for the car to follow, nor does it have dialogues if a command is ambiguous.
In future versions, \texttt{Talk2Car} will be expanded to include the above annotations and dialogues. However, 
this first version already offers a challenging dataset to improve current methods for the joint processing of language and visual data and for the development of suitable machine learning architectures. Especially for cases where the ambiguity in object referral can be resolved by correctly interpreting the constraints found in the language commands, \texttt{Talk2Car} offers a natural and realistic environment to study these. 

\acknowledgments{
This project is sponsored by the MACCHINA project from the KU Leuven  with grant number C14/18/065.
We would like to thank Nvidia for granting us two TITAN Xp GPUs.
We would also like to thank Holger Ceaser from nuTonomy for providing help with the original nuScenes dataset.
We'd also like to thank Tinne Tuytelaars and Matthew Blaschko for their mentorship and good advices.
}

\newpage
\clearpage

\bibliography{emnlp-ijcnlp-2019}
\bibliographystyle{acl_natbib}

\appendix
\newpage
\clearpage
\section{Reproducibility}
\label{sec:supplemental}
In this section we 
disclose the used parameters of some of the mentioned models for reproducibility purposes.
The only models that are not mentioned here are RNM and RS as they just apply a random strategy.

\subsection{OSM}
\label{subsec_appendix: OSM}
We use a SSD-512 model to generate the region proposals. The model was initialized with weights from a model pretrained on ImageNet. We used stochastic gradient descent with initial learning rate 1e-3, momentum 0.9 and weight decay 5e-4 to optimize the loss. The learning rate was degraded by a factor 10 after 80,000 and 100,000 iterations. Batches of size 32 were used during training. Additionally, we found it important to use a warm-up scheme at the start of the training, where we gradually increased the learning rate from 1e-6 to 1e-3, after which we resumed the normal learning rate schedule. We included standard data augmentations during training, i.e., color jitter, random object crops, rescalings and horizontal flips.

The OSM model uses 64 region proposals extracted by the single-shot detection model. We used a ResNet-18 model, pretrained on ImageNet to encode the local regions. A bidirectional GRU model with one hidden layer of size 512 was used to encode the sentence. We optimized the loss with stochastic gradient descent with initial learning rate 1e-3, momentum 0.9 and weight decay 1e-4. We ignored the loss term when there were no region proposals with mean intersection over union larger than 0.5. The learning rate was reduced by a factor 10 when the validation performance became stagnant. We used batches of size 8.

\subsection{SCRC}
We reused the single shot detection model from before (see sec. \ref{subsec_appendix: OSM}) to generate 64 region proposals per image. The local and global features were generated by two separate ResNet-18 models, both pretrained on ImageNet. We used a bidirectional GRU with 512 hidden units for the language model. The local and global recurrent context models use a uni-directional GRU with 512 hidden units.
In the original paper they also pre-train the GRUs on the captioning task. We decided not to do this however as there was no captioning dataset that was close to the dataset described in this paper. These GRUs were thus initialised randomly.
We reused the optimization scheme from section \ref{subsec_appendix: OSM} to train the model. 

\subsection{STACK-NMN}
The images were first resized from $1600 \times 900$ to $512 \times 512$ before extracting the feature maps with ResNet-101.
To extract the feature maps from this model we cut it off at the fourth channel.
The output of the used ResNet model is a tensor of size $32 \times 32 \times 1024$ per image of size $512 \times 512$.
The STACK-NMN model makes use of $Feat_H$ and $Feat_W$ parameters internally representing the amount of feature channels for the height and width. These were both set to $32$.
These parameters are important as they allow the network to transform the center of a ground truth bounding box to a cell in the $32 \times 32$ grid or vice versa.
An other crucial parameter to the STACK-NMN model is the amount of reasoning steps. 
This influences both inference speed as well as accuracy.
We tested the following values: [1, 2, 4, 6, 8, 9, 16] and found that 4 reasoning steps gave us the best results.
The model was trained until the validation accuracy didn't increase over 10 epochs.
The best model on the validation set is saved and used in the experiments.
We also used a batch size of 64.
The rest of the parameters were not changed and were left as the original parameters in the implementation.

\subsection{MAC}
To extract the visual feature maps from the images in the \texttt{Talk2Car} dataset we reuse the method mentioned in STACK-NMN.
We used the following parameters for the MAC model;
We added L2-regularization to the model and used gradient clipping at 5. 
Next, Exponential Moving Average was also used for the weights of the model with a weight decay of 0.999.
We experimented with different learning rates but found that 0.0001 gave us the best results when using the Adam optimizer.
When the loss between two epochs didn't decrease more than $0.2$ we multiplied the learning rate by $0.5$ as is the default value in MAC. 
The following changes were made to transform MAC to the object referral task based on the implementation of STACK-NMN;
We added a cos/sin based positional encoding to the feature maps by concatenation inspired by \cite{vaswani2017attention}.
We also use 32 for $Feat_H$ and $Feat_W$ to calculate the corresponding cell of the center of a bounding box. 
Instead of using the question and memory as the input to the output unit, we used the pre-Softmax attention map from the last read unit of the reasoning process.
This attention map is then passed to a fully connected layer that predicts the cell in which the center of the bounding box lies.
With a convolutional layer we pass over the image to predict the offsets of the bounding box relative to the center of the bounding box.
We also experimented with different amounts of reasoning steps ([1,2,3,4,8,10,12]) and found that with our modified version of MAC 10 reasoning steps worked the best for our task. 
The batch size was set to 32.
The rest of the parameters remained unchanged to the original paper.

\subsection{BOBB}
\begin{figure*}[t]
\begin{center}
 \centering
\subfloat[Training set]{\includegraphics[width=.32\linewidth]{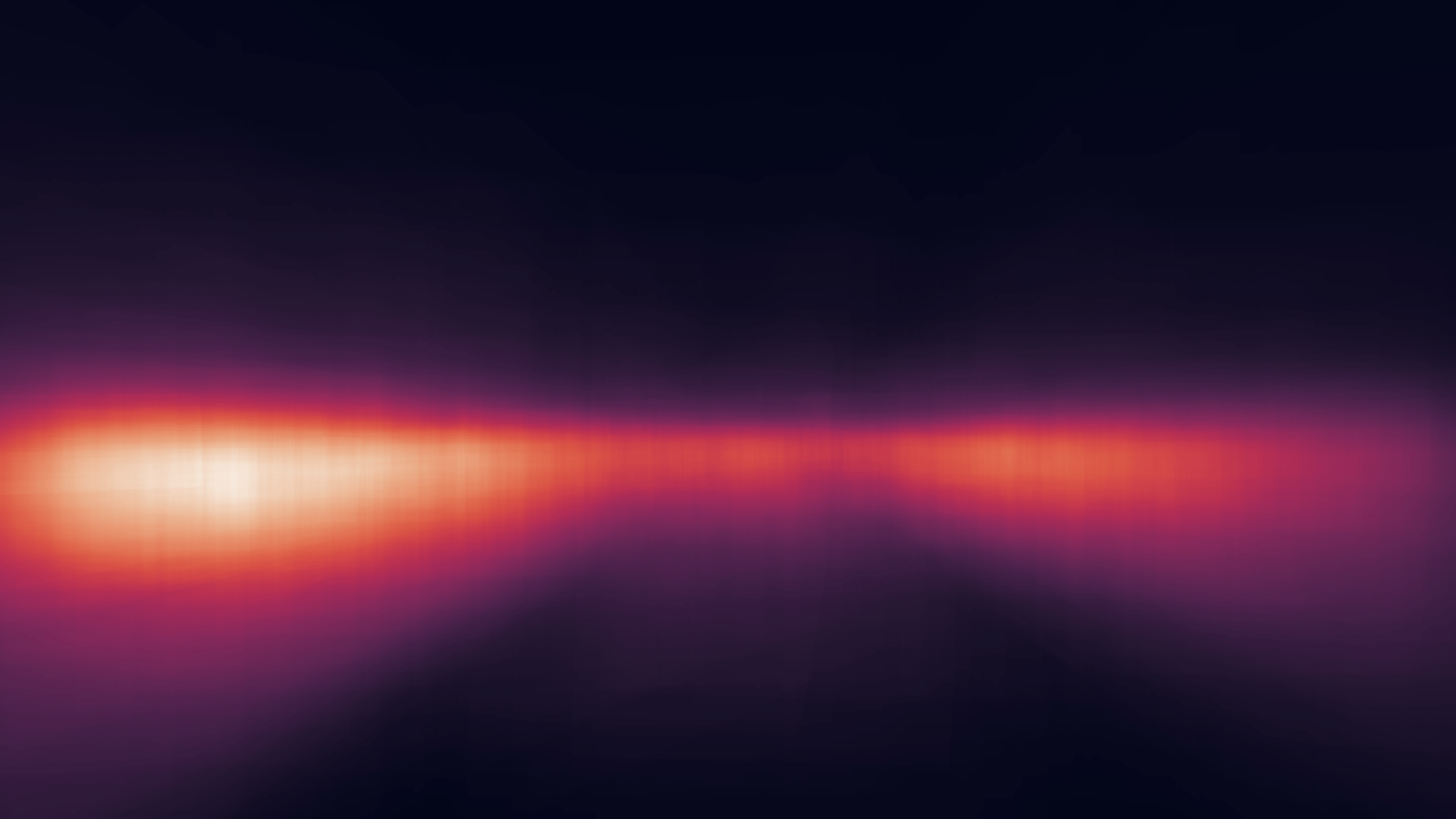}}
\hfill
\subfloat[Validation set]{\includegraphics[width=.32\linewidth]{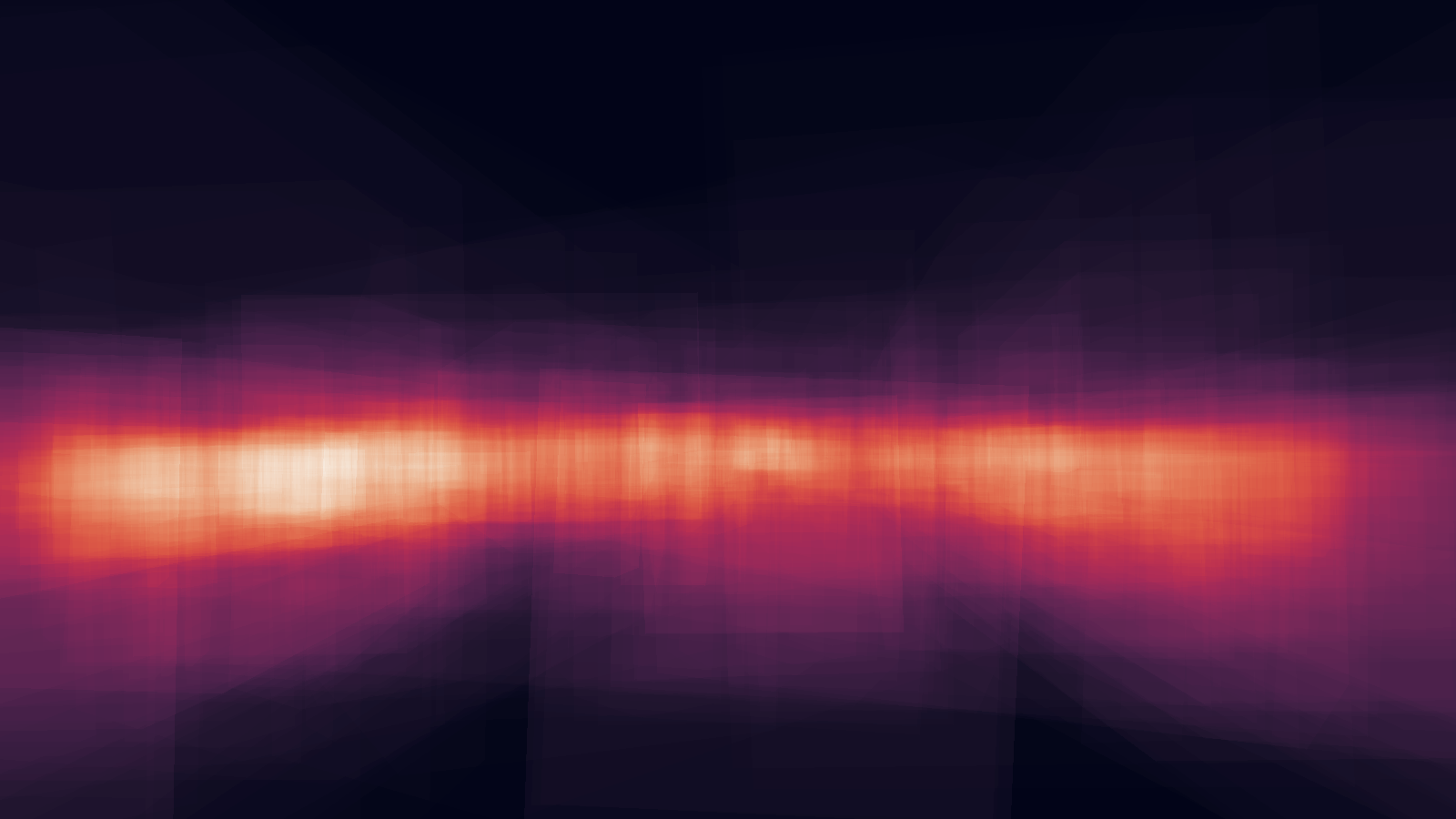}}
    \hfill
\subfloat[Test set]{\includegraphics[width=.32\linewidth]{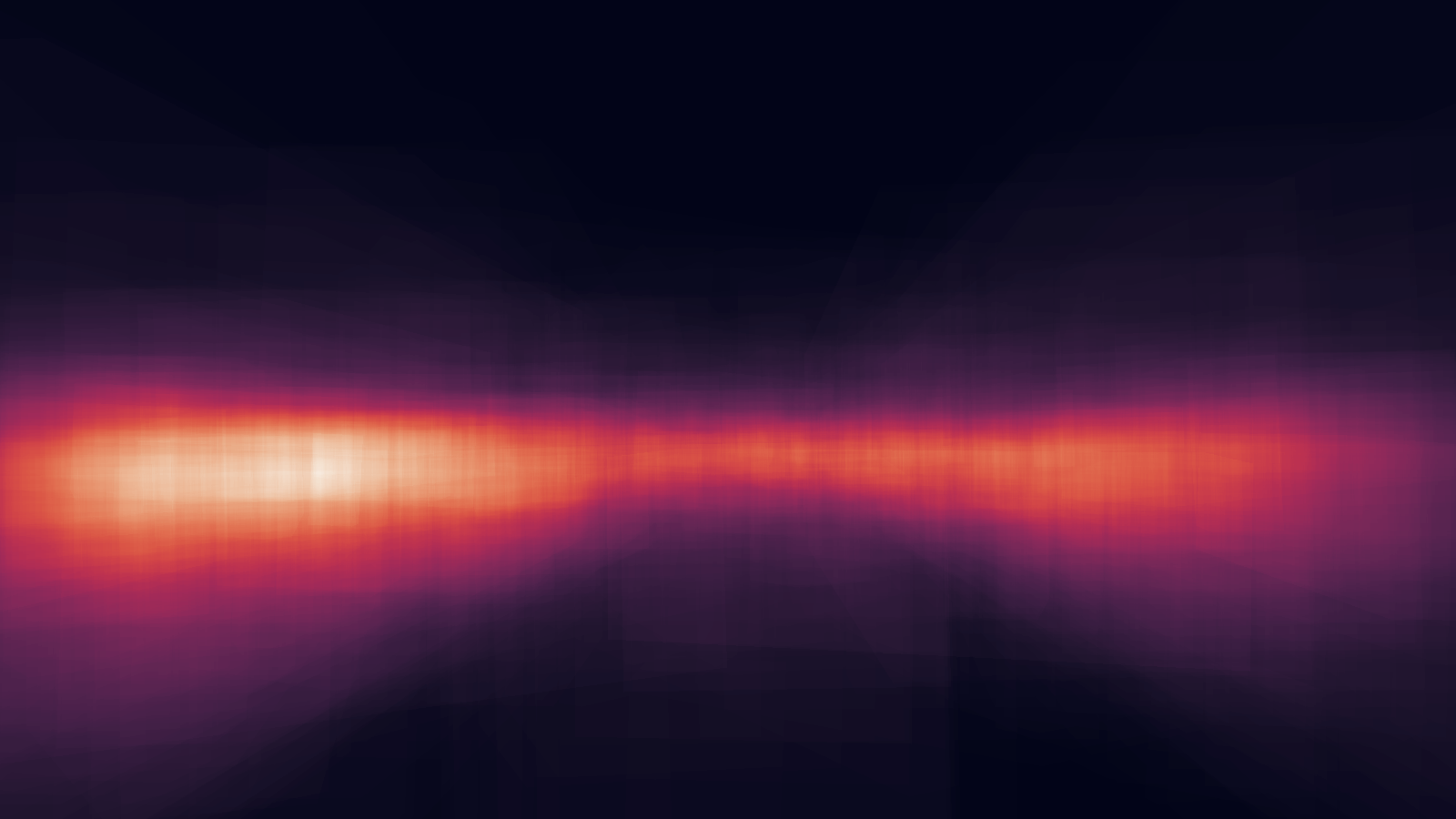}}
\end{center}
\caption{The heatmaps of the locations of all objects in the training set (a), validation set (b) and test set (c) respectively.}
\label{fig:heatmaps}
\end{figure*}

The algorithm that is used for this model is described in Algorithm \ref{algo:BOBB} in section \ref{sec:supplemental} and has been used on the bounding boxes of the training set.
A heatmap of the location of the objects in the training set can be seen in Fig. \ref{fig:heatmaps}(a).
The resulting bounding box that was found is:  
[0, 435, 445, 325]
with format $[x_1, y_1, w, h]$.
$x_1$ and $y_1$ represent the lower left corner of the bounding box.
This found bounding box corresponds with the bias seen on the map.


\begin{algorithm*}
\caption{BOBB Algorithm}\label{algo:BOBB}
\begin{algorithmic}[1]
\Procedure{FindBestBBox}{train\_gt\_bboxes, imgWidth=1600, imgHeight=900, threshold=0.5}
\State $bestAm \gets 0$
\State $bestBox \gets None$
\State $X \gets \text{linspace}(0, \text{ImgWidth}, \text{step}=5)$
\State $Y \gets \text{linspace}(0, \text{ImgHeight}, \text{step}=5)$
\For{$x_1$ in $X$}{
    \For{$x_2$ in $X$}{
    \If {$x_2 \leq x_1$} continue
    \EndIf
    \For{$y_1$ in $Y$}{
    \For{$y_2$ in $Y$}{
    \If {$y_2 \leq y_1$} continue
    \EndIf
    \State $box \gets [x_1,y_1,x_2-x_1, y_2-y_1]$
    \State $am \gets getAmountOfIoUAboveThresh(box, train\_gt\_bboxes, threshold)$
    \If {$am > bestAm$}
    \State $bestAm \gets am$
    \State $bestBox \gets box$
    \EndIf
    }
    \EndFor
    }
    \EndFor
    }
    \EndFor
}
\EndFor
\State \Return $bestBox$
\EndProcedure
\end{algorithmic}
\end{algorithm*}

\end{document}